\documentclass{article} % For LaTeX2e
\usepackage{iclr2025_conference,times}

\usepackage{amsmath,amsfonts,bm}

\def\eqref#1{equation~\ref{#1}}
\def\1{\bm{1}}

\DeclareMathAlphabet{\mathsfit}{\encodingdefault}{\sfdefault}{m}{sl}
\SetMathAlphabet{\mathsfit}{bold}{\encodingdefault}{\sfdefault}{bx}{n}

\usepackage{amsmath,amssymb,amsfonts}
\usepackage{algorithmic}
\usepackage{graphicx}
\usepackage{textcomp}
\usepackage{xcolor}
\usepackage{booktabs}
\usepackage{multirow}
\usepackage{hyperref}
\usepackage{url}

\title{Support–Contra Asymmetry in LLM Explanations}

\iclrfinalcopy

\author{Avinash Patil \\
Hewlett Packard Enterprise \\
Sunnyvale, USA \\
\texttt{avinash.patil@hpe.com} \\
\texttt{ORCID: 0009-0002-6004-370X}
}

\begin{document}

\maketitle
\begin{abstract}

Large Language Models (LLMs) increasingly produce natural language explanations alongside their predictions, yet it remains unclear whether these explanations reference predictive cues present in the input text. In this work, we present an empirical study of how LLM-generated explanations align with predictive lexical evidence from an external model in text classification tasks. To analyze this relationship, we compare explanation content against interpretable feature importance signals extracted from transparent linear classifiers. These reference models allow us to partition predictive lexical cues into \emph{supporting} and \emph{contradicting} evidence relative to the predicted label.

Across three benchmark datasets—\textsc{WikiOntology}, \textsc{AG News}, and \textsc{IMDB}—we observe a consistent empirical pattern that we term \textbf{support–contra asymmetry}. Explanations accompanying correct predictions tend to reference more supporting lexical cues and fewer contradicting cues, whereas explanations associated with incorrect predictions reference substantially more contradicting evidence. This pattern appears consistently across datasets, across reference model families (logistic regression and linear SVM), and across multiple feature retrieval depths.

These results suggest that LLM explanations often reflect lexical signals that are predictive for the task when predictions are correct, while incorrect predictions are more frequently associated with explanations that reference misleading cues present in the input. Our findings provide a simple empirical perspective on explanation–evidence alignment and illustrate how external sources of predictive evidence can be used to analyze the behavior of LLM-generated explanations.

\end{abstract}

\section{Introduction}

Large Language Models (LLMs) explanations—often presented as rationales accompanying model outputs—play an important role in explainable AI systems by helping users interpret model behavior in applications such as sentiment analysis, topic classification, and information retrieval. However, a central question remains: \emph{do these explanations reference predictive cues present in the input text?}

Answering this question is challenging because LLM explanations are free-form natural language. Unlike traditional machine learning models, where feature importance scores and attribution methods provide interpretable signals about which input features influence a prediction, LLM explanations may appear plausible even when they do not correspond to measurable predictive evidence in the input. As a result, explanations can be fluent and convincing while remaining weakly connected to the cues that drive model predictions.

In this work, we present an empirical study of how LLM-generated explanations relate to predictive lexical evidence in text classification tasks. Our analysis reveals a consistent empirical pattern that we term \textbf{support–contra asymmetry}. Intuitively, if explanations reference predictive cues in the input, they should emphasize evidence that supports the predicted label while avoiding cues that contradict it. Across multiple datasets, we observe that explanations accompanying correct predictions tend to reference more supporting evidence and less contradicting evidence, whereas explanations associated with incorrect predictions reference substantially more contradicting lexical cues.

To analyze this phenomenon, we compare explanation content against predictive lexical signals extracted from transparent linear classifiers. These classifiers serve as \emph{reference evidence extractors}: they identify lexical features that are strongly associated with particular labels in the training data. For each instance, we partition these predictive cues into two categories: \emph{supporting features}, which increase the likelihood of the predicted label, and \emph{contradicting features}, which provide evidence against it. Measuring whether explanations reference these cues provides a quantitative view of \emph{explanation–evidence alignment}.

We evaluate this approach on three benchmark datasets—\textsc{WikiOntology}, \textsc{AG News}, and \textsc{IMDB}—covering both topical classification and sentiment analysis tasks. Across datasets, reference models, and feature retrieval depths, we consistently observe support–contra asymmetry: explanations for correct predictions exhibit higher coverage of supporting features and lower coverage of contradicting features, while explanations associated with incorrect predictions reference substantially more contradicting evidence. This pattern is robust across reference model families (logistic regression and linear SVM) and across multiple feature depths.

These findings provide new insight into the behavior of LLM explanations. When predictions are correct, explanations frequently reference lexical signals that align with predictive evidence identified by interpretable reference models. When predictions are incorrect, explanations more often highlight misleading cues present in the input. This asymmetry suggests that explanation content tends to track the direction of predictive lexical evidence present in the input text as identified by the reference classifiers, rather than functioning purely as a post-hoc narrative.

The contributions of this work are as follows:

\begin{itemize}

\item We identify and empirically characterize \textbf{support–contra asymmetry} in LLM-generated explanations, a consistent pattern in which explanations accompanying correct predictions emphasize supporting lexical evidence while explanations for incorrect predictions reference more contradicting cues from the input text.

\item We introduce a simple quantitative framework for measuring explanation–evidence alignment by comparing explanation text with predictive lexical signals extracted from transparent reference classifiers.

\item We demonstrate that the observed asymmetry replicates across multiple datasets, reference model families, and feature retrieval depths.

\item We provide empirical evidence that LLM explanations often reflect predictive lexical signals present in the input while also revealing systematic shifts in explanation behavior when predictions are incorrect.

\end{itemize}

The remainder of this paper is organized as follows. Section~\ref{related_work} reviews prior work on explanation faithfulness and interpretable models. Section~\ref{methodology} describes the measurement framework and alignment metrics. Section~\ref{experiments} presents the experimental setup. Section~\ref{results} reports the empirical findings, followed by Section~\ref{discussion}, which discusses implications and limitations. Section~\ref{conclusion} concludes the paper.

\section{Related Work}
\label{related_work}

\paragraph{Faithfulness in NLP explanations.}
A central distinction in explainable NLP is between \emph{plausibility}---whether an explanation appears convincing to humans---and \emph{faithfulness}---whether it accurately reflects the model's decision process. Jacovi and Goldberg~\cite{jacovi-goldberg-2020-towards} argue that these notions should be separated explicitly, and Lyu et al.~\cite{lyu-etal-2024-towards} provide a broad survey of faithful explanation methods in NLP. In text classification, prior work has also emphasized that explanation methods should be assessed not only by readability but also by whether they track predictive behavior under perturbation, sufficiency, and comprehensiveness tests~\cite{deyoung-etal-2020-eraser,atanasova-etal-2020-diagnostic}.

\paragraph{Natural language explanations and rationales.}
Natural language explanations (NLEs), often called \emph{rationales}, are appealing because they are directly legible to users. The ERASER benchmark~\cite{deyoung-etal-2020-eraser} established a widely used framework for evaluating rationale-based NLP models by combining task labels with human evidence annotations. At the same time, work on explainability for text classification has shown that rationale quality can differ substantially across explanation methods and architectures~\cite{atanasova-etal-2020-diagnostic}. More recent work has highlighted that evaluating free-text explanations is especially difficult because they are unconstrained, can be semantically diffuse, and may not map cleanly onto measurable evidence in the input~\cite{atanasova-etal-2023-faithfulness,parcalabescu-frank-2024-measuring}.

\paragraph{Evaluating faithfulness of free-text explanations.}
A growing line of work proposes direct tests for whether free-text explanations are behaviorally faithful. Atanasova et al.~\cite{atanasova-etal-2023-faithfulness} introduce counterfactual and reconstruction-based tests for natural language explanations, showing that explanations can remain plausible while failing to reflect decisive evidence. Parcalabescu and Frank~\cite{parcalabescu-frank-2024-measuring} further argue that many current evaluations capture forms of self-consistency rather than full faithfulness, clarifying the methodological difficulty of verifying explanation quality when the model's internal decision process is latent. This concern is echoed by Agarwal et al.~\cite{agarwal-etal-2024-faithfulness}, who emphasize the gap between plausibility and faithfulness in LLM self-explanations.

\paragraph{LLM self-explanations and chain-of-thought.}
The rise of instruction-tuned LLMs has intensified interest in \emph{self-explanations}: free-text explanations produced by the same model that makes the prediction. However, recent work suggests that such explanations should not be assumed to be faithful. Turpin et al.~\cite{turpin-etal-2023-language} show that chain-of-thought explanations can omit influential biasing factors and rationalize incorrect answers. Madsen et al.~\cite{madsen-etal-2024-self} study several explanation formats for LLMs and find that faithfulness is highly explanation-, model-, and task-dependent. Related work on counterfactual simulatability also finds that LLM explanations often fail to support accurate mental models of model behavior on nearby inputs~\cite{chen-etal-2024-do,dehghanighobadi-etal-2025-llms}. More recent studies investigate consistency across related examples~\cite{chen-etal-2025-towards}, the faithfulness of intermediate reasoning traces~\cite{tutek-etal-2025-measuring}, and methods for refining explanations through explicit critique or reward design~\cite{li-etal-2025-drift,wang-atanasova-2025-self,zhao-iii-2025-necessary}.

\paragraph{External evidence and behavior-based auditing.}
Our work is most closely related to approaches that evaluate explanations against \emph{external} evidence rather than relying on the model's self-report alone. Existing work has used human rationales~\cite{deyoung-etal-2020-eraser}, perturbation tests~\cite{atanasova-etal-2023-faithfulness}, or simulatability under counterfactuals~\cite{chen-etal-2024-do} as outside criteria. Recent work also studies whether faithfulness metrics themselves are diagnostic of genuine explanation quality~\cite{zaman-srivastava-2025-causal}. Our approach differs in using transparent linear classifiers as reference evidence extractors for text classification. Rather than asking whether an explanation is globally faithful to the full internal computation of an LLM, we ask a narrower and empirically tractable question: does the explanation mention lexical cues that are externally predictive for the task? This framing lets us compare explanation content against signed lexical evidence and separate such evidence into cues that support versus contradict the predicted label.

\paragraph{Positioning of the present work.}
Relative to prior work, our contribution is not a new explanation-generation method or a new benchmark for human plausibility. Instead, we provide a simple empirical lens on explanation--evidence alignment in text classification. By partitioning externally predictive lexical features into \emph{supporting} and \emph{contradicting} evidence, we uncover a robust pattern---\textbf{support--contra asymmetry}---across datasets, models, and reference classifiers. This complements prior studies on rationale faithfulness by showing that even a lightweight outside model can reveal systematic differences between explanations attached to correct versus incorrect LLM predictions.

\section{Methodology}
\label{methodology}

This section describes the framework used to measure alignment between LLM-generated explanations and predictive lexical cues identified by transparent linear classifiers. The framework quantifies how frequently lexical features that support or contradict a predicted label—according to the reference classifier—appear in the explanation produced by the LLM. The goal is to analyze whether explanations preferentially reference cues associated with the predicted label and avoid cues that oppose it, and whether this pattern differs when predictions are incorrect.

Importantly, predictive features are derived from external reference classifiers rather than the LLM itself. Consequently, the resulting metrics measure alignment with an \emph{external lexical evidence baseline} rather than direct faithfulness to the LLM’s internal reasoning process. Overlap between explanations and classifier-derived features should therefore be interpreted as correspondence with predictive lexical cues identified by these reference models rather than as a direct probe of the LLM’s internal decision process.

\subsection{Problem Formulation}
\label{problem-formulation}

Let the dataset be $\mathcal{D} = \{(x_i, y_i)\}_{i=1}^N$, where $x_i$ is a text instance and $y_i$ is its ground-truth label.  
For each instance, an LLM produces a predicted label $\hat{y}_i$ and an accompanying natural-language explanation $e_i$.

To obtain transparent feature-level evidence, we train linear text classifiers using TF--IDF features. Two model families are used:

\begin{itemize}
    \item Logistic Regression
    \item Linear Support Vector Machines
\end{itemize}

Because these models are linear, the influence of each vocabulary feature on a prediction can be directly inspected via its coefficient weight. For each instance $x_i$, we extract the top-$k$ most influential lexical features associated with the predicted class:

\[
F_i = \{(f_{i1}, w_{i1}), \dots, (f_{ik}, w_{ik})\},
\]

where $f_{ij}$ denotes a lexical feature and $w_{ij}$ its associated model weight.

We partition these features into supporting and contradicting evidence:

\[
\mathcal{S}_i = \{f_{ij} : w_{ij} > 0\}, \quad
\mathcal{C}_i = \{f_{ij} : w_{ij} < 0\}.
\]

Supporting features increase the likelihood of the predicted class under the reference classifier, whereas contradicting features provide evidence against it.

\subsection{Feature--Explanation Matching}

To determine whether a feature $f$ appears in an explanation $e_i$, we apply lexical matching functions. Two matching strategies are used:

\begin{itemize}
    \item \textbf{Token-aware matching:} Features are matched after normalization including lowercasing, punctuation removal, and lemmatization.
    \item \textbf{Exact matching:} Strict string equality between the feature and any span in the explanation.
\end{itemize}

Formally, for matcher type $\tau \in \{\text{token}, \text{exact}\}$:

\[
m_\tau(f, e_i) =
\begin{cases}
1 & \text{if feature } f \text{ appears in } e_i \\
0 & \text{otherwise.}
\end{cases}
\]

For token-aware matching, both explanation text and candidate features are normalized by lowercasing, punctuation stripping, and lemmatization. Unigram features are matched against normalized explanation tokens. Bigram features are matched against contiguous normalized two-token spans in the explanation.

\subsection{Coverage Metrics}

We measure how often supporting and contradicting features appear in explanations. For a subset $\mathcal{T} \in \{\mathcal{S}_i, \mathcal{C}_i\}$ the per-instance coverage score is

\[
\text{cov}_\tau(i, \mathcal{T}) =
\frac{1}{|\mathcal{T}|}
\sum_{f \in \mathcal{T}} m_\tau(f, e_i).
\]

This yields two primary metrics:

\[
\text{support\_cov}_{i,\tau} = \text{cov}_\tau(i, \mathcal{S}_i),
\quad
\text{contra\_cov}_{i,\tau} = \text{cov}_\tau(i, \mathcal{C}_i).
\]

These values represent the proportion of supporting or contradicting reference-model features that appear in the LLM explanation.

\subsection{Correctness Partition}

To analyze whether explanation–evidence alignment differs between correct and incorrect predictions, instances are partitioned according to prediction correctness:

\[
\text{match}_i = \mathbf{1}[\hat{y}_i = y_i],
\quad
\text{mismatch}_i = \mathbf{1}[\hat{y}_i \neq y_i].
\]

Coverage statistics are aggregated separately for these two subsets.

We define the difference in mean coverage between the two groups:

\[
\Delta_{\text{support}} =
\mathbb{E}[\text{support\_cov} \mid \text{match}]
-
\mathbb{E}[\text{support\_cov} \mid \text{mismatch}]
\]

\[
\Delta_{\text{contra}} =
\mathbb{E}[\text{contra\_cov} \mid \text{match}]
-
\mathbb{E}[\text{contra\_cov} \mid \text{mismatch}]
\]

To quantify the relative imbalance between supporting and contradicting evidence referenced by explanations, we define the asymmetry metric:

\[
\Delta^* = \Delta_{\text{support}} - \Delta_{\text{contra}}.
\]

A positive $\Delta^*$ indicates that explanations emphasize supporting evidence while downplaying contradicting evidence when predictions are correct.

\subsection{Statistical Inference}

Statistical significance is estimated using bootstrap resampling. For each dataset and model configuration, we repeatedly resample instances with replacement and recompute the coverage differences.

From the bootstrap distribution we compute mean difference estimates
and 95\% confidence intervals for all coverage statistics.

Bootstrap inference is applied to $\Delta_{\text{support}}$, $\Delta_{\text{contra}}$, and the asymmetry statistic $\Delta^*$.

\subsection{Experimental Settings}

Experiments are conducted across multiple datasets representing different text classification tasks:

\begin{itemize}
    \item IMDB sentiment classification
    \item AG News topic classification
    \item WikiOntology entity classification
\end{itemize}

For each instance, the top-$k$ most influential features are extracted from the linear model, with

\[
k \in \{3,5,7\}.
\]

Evaluating multiple $k$ values allows us to examine whether alignment between explanations and model evidence persists beyond the most dominant features.

\subsection{Implementation Details}

Text classifiers are implemented using scikit-learn with TF--IDF representations including unigrams and bigrams. Standard preprocessing steps include lowercasing, stopword removal, and vocabulary filtering.

Matching operations and coverage metrics are implemented in Python. All experiments are executed using fixed dataset splits and consistent preprocessing pipelines to ensure reproducibility.

\section{Experiments}
\label{experiments}

This section describes the experimental setup used to evaluate the relationship between LLM explanations and classifier-derived feature evidence. We introduce the datasets, the procedure for generating LLM predictions and explanations, the reference models used to extract lexical evidence, and the statistical evaluation protocol.

\subsection{Datasets}

Experiments are conducted on three benchmark text classification datasets covering different task types and label granularities. Experiments were conducted on the standard test sets of the IMDB, AG News, and WikiOntology dataset. We derived “Ground truth” feature importances from logistic regression and linear SVM from the same dataset and label semantics as those used to evaluate the LLM. Training LR on a different dataset may shift feature weights (e.g., words signaling “positive” in one corpus may be neutral or rare in another), this makes the comparison less about LLM alignment with model features, and more about dataset/domain mismatch.

\begin{itemize}
    \item \textbf{WikiOntology:} A multi-class entity classification dataset consisting of short textual descriptions mapped to ontology categories (e.g., \emph{Album}, \emph{Company}, \emph{Film}). The dataset contains 15 entity classes.
    
    \item \textbf{AG News:} A four-class news topic classification dataset with categories \emph{World}, \emph{Sports}, \emph{Business}, and \emph{Sci/Tech}. Instances consist of short news headlines and summaries.
    
    \item \textbf{IMDB:} A binary sentiment classification dataset containing movie reviews labeled as \emph{Positive} or \emph{Negative}.
\end{itemize}

These datasets span different classification regimes (binary vs.\ multi-class) and domains (sentiment, news topics, and entity categorization), allowing us to evaluate explanation--evidence alignment across heterogeneous tasks.

\subsection{LLM Predictions and Explanations}

All predictions and explanations were generated using the DeepSeek-R1 large
language model. For each input instance, the model was prompted to output both
a predicted label and a natural language explanation. Decoding used deterministic generation (temperature = 0, top-p = 1) to ensure reproducibility. For each instance, the LLM is prompted to output both a predicted label and a natural-language explanation describing the reasoning behind the prediction. The explanation text serves as the rationale $e_i$ defined in Section~\ref{methodology}. A total of 25,000 IMDB reviews, 7,458 AG News articles, and 70,000 WikiOntology instances were evaluated. The LLM achieved accuracies of 96.5\%, 84.7\%, and 96.9\% on the IMDB, AG News, and WikiOntology datasets respectively. Prompt templates for all 3 datasets are attached in appendix.

Prediction correctness is determined by comparing the predicted label $\hat{y}_i$ with the ground-truth label $y_i$. Instances are partitioned into two groups:

\begin{itemize}
    \item \textbf{Match:} $\hat{y}_i = y_i$
    \item \textbf{Mismatch:} $\hat{y}_i \neq y_i$
\end{itemize}

This partition allows us to analyze whether explanations reference different types of evidence when the prediction is correct versus when it is incorrect.

\subsection{Reference Feature Models}

To obtain interpretable lexical evidence, we train transparent linear classifiers using TF--IDF features. Two model families are used:

\begin{itemize}
    \item Logistic Regression
    \item Linear Support Vector Machines (SVM)
\end{itemize}

Because these models are linear, the contribution of each vocabulary feature to a prediction can be directly inspected through its coefficient weight.

For each instance, we extracted top-$k$ supporting and contradicting features from the reference classifier using only features that were active in that instance, i.e., features with nonzero TF--IDF value in the document. Feature importance was computed at the instance level rather than from raw classifier coefficients alone: for a document $x$ and the classifier’s predicted class $c$, each feature $f$ was assigned a signed contribution score equal to $\mathrm{TFIDF}(f,x) \times \beta_{c,f}$, where $\beta_{c,f}$ is the class-specific logistic regression coefficient. Supporting features were defined as features with positive contribution scores, while contradicting features were defined as features with negative contribution scores relative to the predicted class. We then ranked supporting features by decreasing positive contribution and contradicting features by increasing negative contribution magnitude, and retained the top-$k$ features from each set for downstream coverage analysis.

\begin{itemize}
    \item \textbf{Supporting features:} $w_{ij} > 0$, which increase the likelihood of the predicted label.
    \item \textbf{Contradicting features:} $w_{ij} < 0$, which provide evidence against the predicted label.
\end{itemize}

This decomposition allows explanations to be compared against both positive and negative evidence used by the classifier.

\subsection{Feature Depth Analysis}

To test whether explanation alignment depends on the number of retrieved classifier features, we vary the feature depth parameter:

\[
k \in \{3,5,7\}.
\]

Smaller values focus on the strongest predictive signals, while larger values include weaker features. Consistency across these settings indicates that the observed explanation patterns do not depend on a particular feature cutoff.

\subsection{Feature Masking Validation}
To validate that the lexical cues extracted from the reference model capture behaviorally relevant signals for the LLM, we conduct a feature ablation study on AG News. For each input, we remove the top supporting features, contradicting features, or a random set of features and re-evaluate the model. Removing supporting features slightly reduces accuracy relative to random removal (0.8475 vs. 0.8537), while removing contradicting features improves performance (0.8768), suggesting that contradiction cues introduce conflicting evidence for the model. At the instance level, approximately 7.3\% of predictions change when support features are removed instead of random features, indicating sensitivity to these cues. Directional analysis further shows that removing support features breaks correct predictions more often than it fixes errors (276 vs. 230), whereas removing contradicting features corrects substantially more errors than it introduces (270 vs. 98). Together, these results indicate that the lexical cues identified by the reference model capture behaviorally meaningful signals for the LLM, supporting their use as a diagnostic baseline for evaluating explanation–evidence alignment.

\subsection{Coverage Metrics}

Explanation alignment is measured using the coverage metrics defined in Section~\ref{methodology}. For each instance we compute:

\begin{itemize}
    \item Supporting feature coverage
    \item Contradicting feature coverage
\end{itemize}

Coverage is computed using two lexical matching strategies:

\begin{itemize}
    \item Token-aware matching
    \item Exact string matching
\end{itemize}

Dataset-level averages are then computed separately for match and mismatch subsets.

\subsection{Visualization of Explanation--Evidence Alignment}

To visualize how explanations reference different types of evidence, we generate scatter plots where each instance is represented by a point in a two-dimensional space:

\begin{itemize}
    \item $x$-axis: contradicting feature coverage
    \item $y$-axis: supporting feature coverage
\end{itemize}

Points are color-coded according to prediction correctness (match vs.\ mismatch). These plots illustrate how explanations distribute across supporting and contradicting evidence and provide an intuitive visualization of the asymmetry between correct and incorrect predictions.

In particular, explanations associated with correct predictions tend to cluster in regions with higher supporting coverage and lower contradicting coverage, whereas incorrect predictions exhibit comparatively higher contradicting coverage.

\subsection{Statistical Analysis}

To quantify differences between correct and incorrect predictions, we compute three statistics:

\begin{itemize}
    \item $\Delta_{\text{support}}$: difference in support coverage between matches and mismatches
    \item $\Delta_{\text{contra}}$: difference in contradicting coverage between matches and mismatches
    \item $\Delta^*$: the asymmetry statistic defined as $\Delta_{\text{support}} - \Delta_{\text{contra}}$
\end{itemize}

Statistical significance is estimated using bootstrap resampling over instances. For each dataset, model type, feature depth, and matcher, we compute:

\begin{itemize}
    \item mean coverage differences
    \item 95\% bootstrap confidence intervals
\end{itemize}

This analysis allows us to determine whether LLM explanations systematically emphasize supporting evidence and suppress contradicting evidence when predictions are correct.

\section{Results}
\label{results}

We analyze how LLM-generated explanations align with predictive lexical features extracted from transparent linear classifiers across three datasets: \textsc{WikiOntology}, \textsc{AG News}, and \textsc{IMDB}. Our analysis focuses on coverage-oriented metrics that measure how frequently explanations reference \emph{supporting} and \emph{contradicting} classifier features.

Statistical reliability is assessed using bootstrap confidence intervals; intervals that exclude zero indicate reliable differences between match and mismatch conditions.

\subsection{Support and Contradiction Coverage}

Coverage patterns reveal a consistent empirical asymmetry between correct and incorrect predictions. When the LLM prediction matches the true label, explanations exhibit higher coverage of \emph{supporting features}. In contrast, explanations accompanying incorrect predictions show elevated coverage of \emph{contradicting features}. This pattern corresponds to the \textbf{support–contra asymmetry} introduced earlier.

\begin{figure}[t]
\centering
\includegraphics[width=0.85\linewidth]{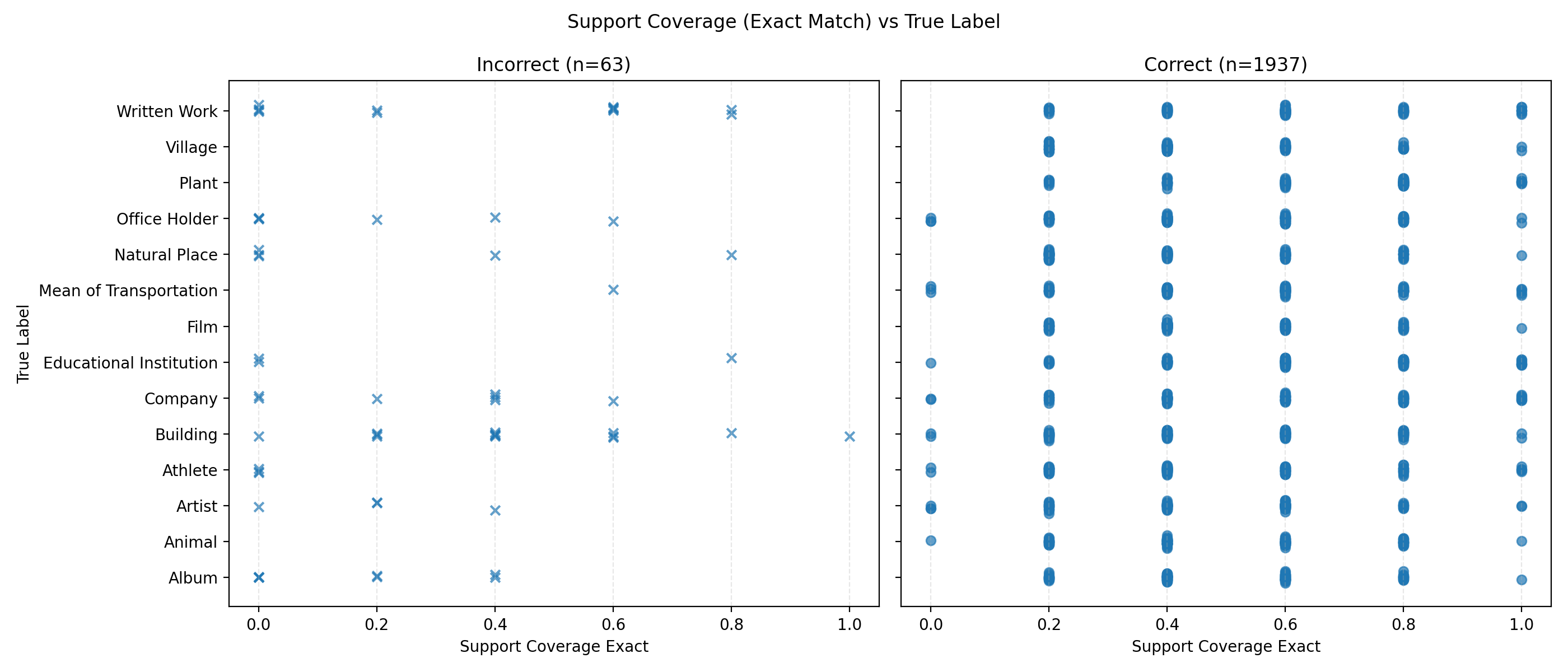}
\includegraphics[width=0.85\linewidth]{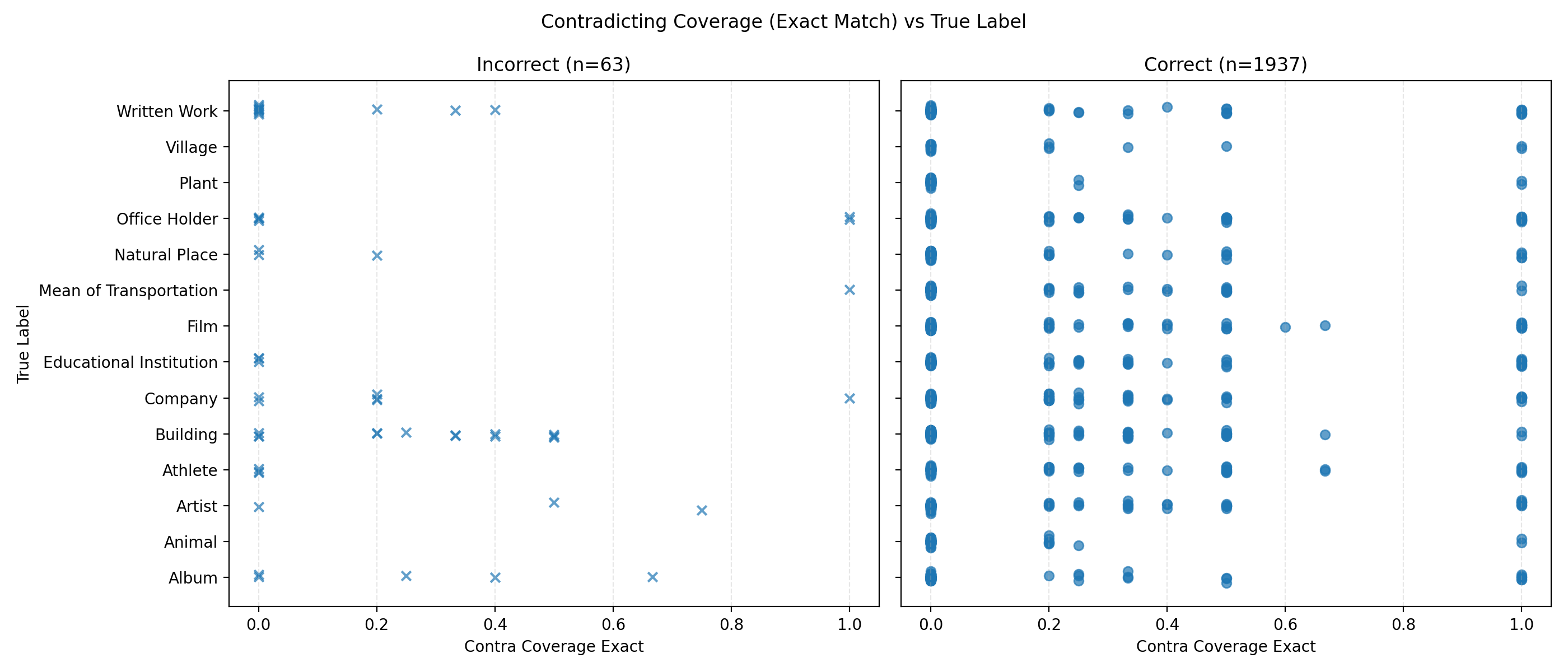}
\caption{Scatter plots of support and contradiction coverage for \textsc{WikiOntology}.}
\label{fig:wiki-scatter}
\end{figure}

\begin{figure}[t]
\centering
\includegraphics[width=0.85\linewidth]{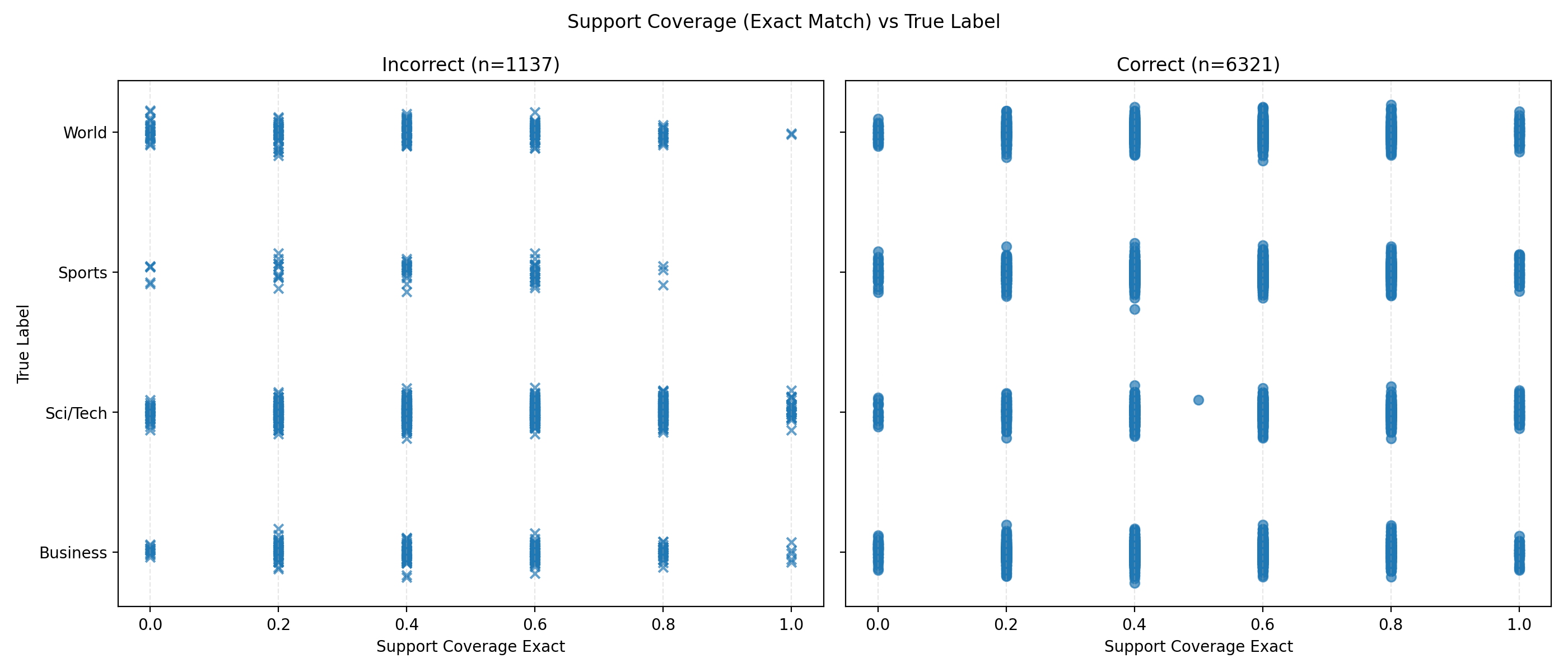}
\includegraphics[width=0.85\linewidth]{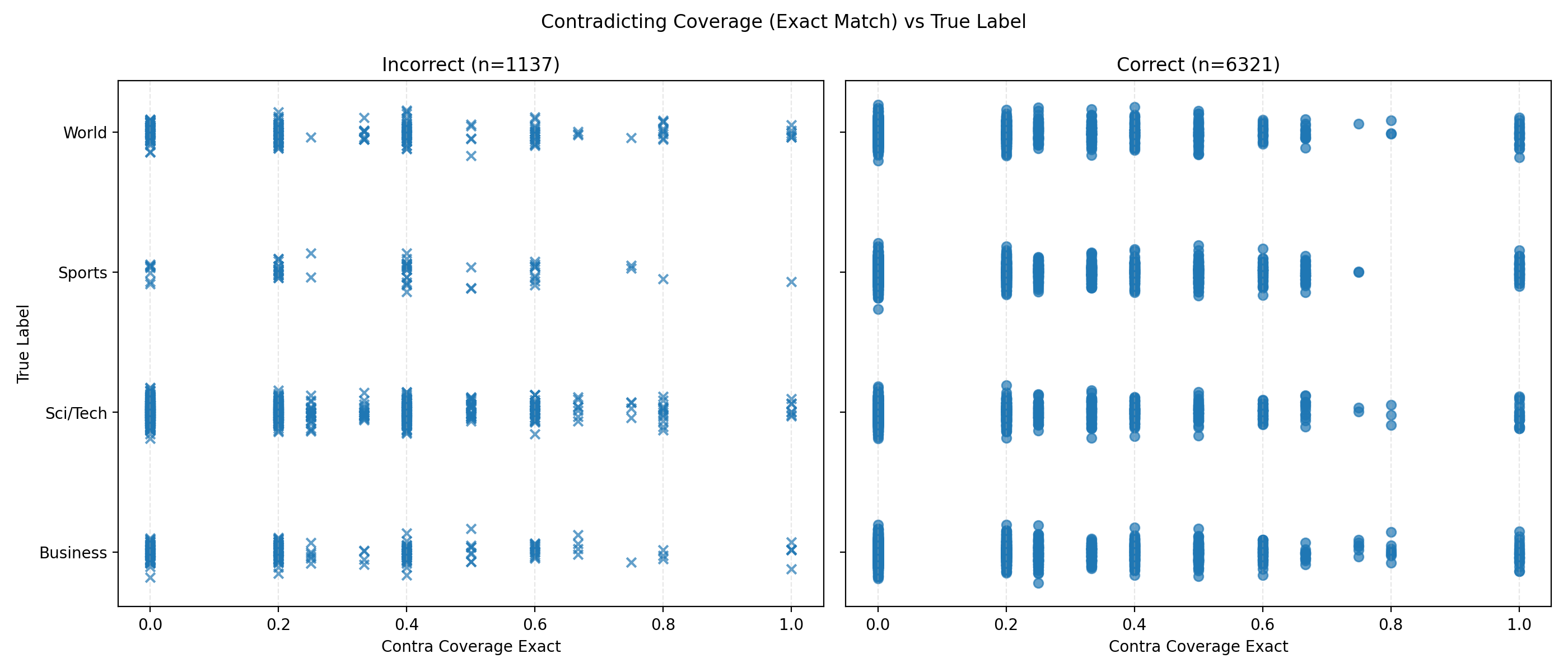}
\caption{Scatter plots of support and contradiction coverage for \textsc{AG News}.}
\label{fig:agnews-scatter}
\end{figure}

\begin{figure}[t]
\centering
\includegraphics[width=0.85\linewidth]{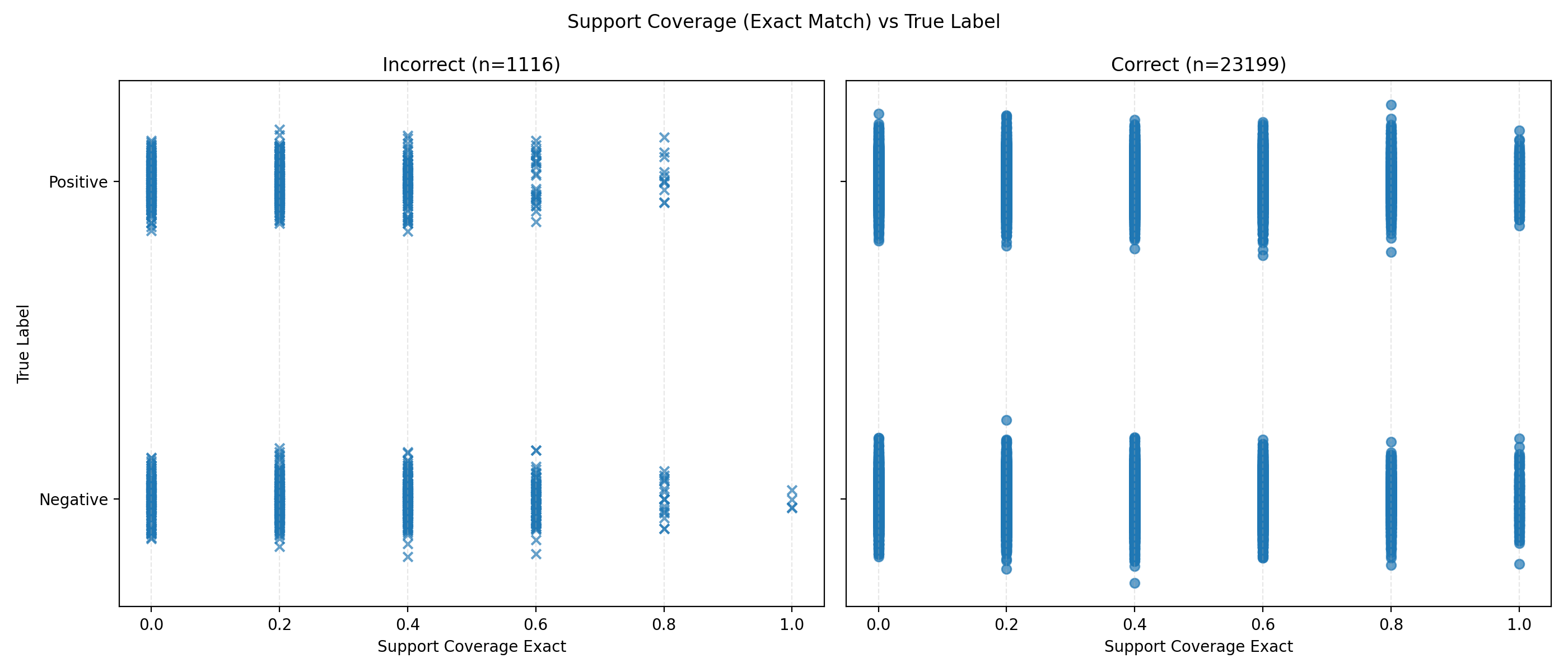}
\includegraphics[width=0.85\linewidth]{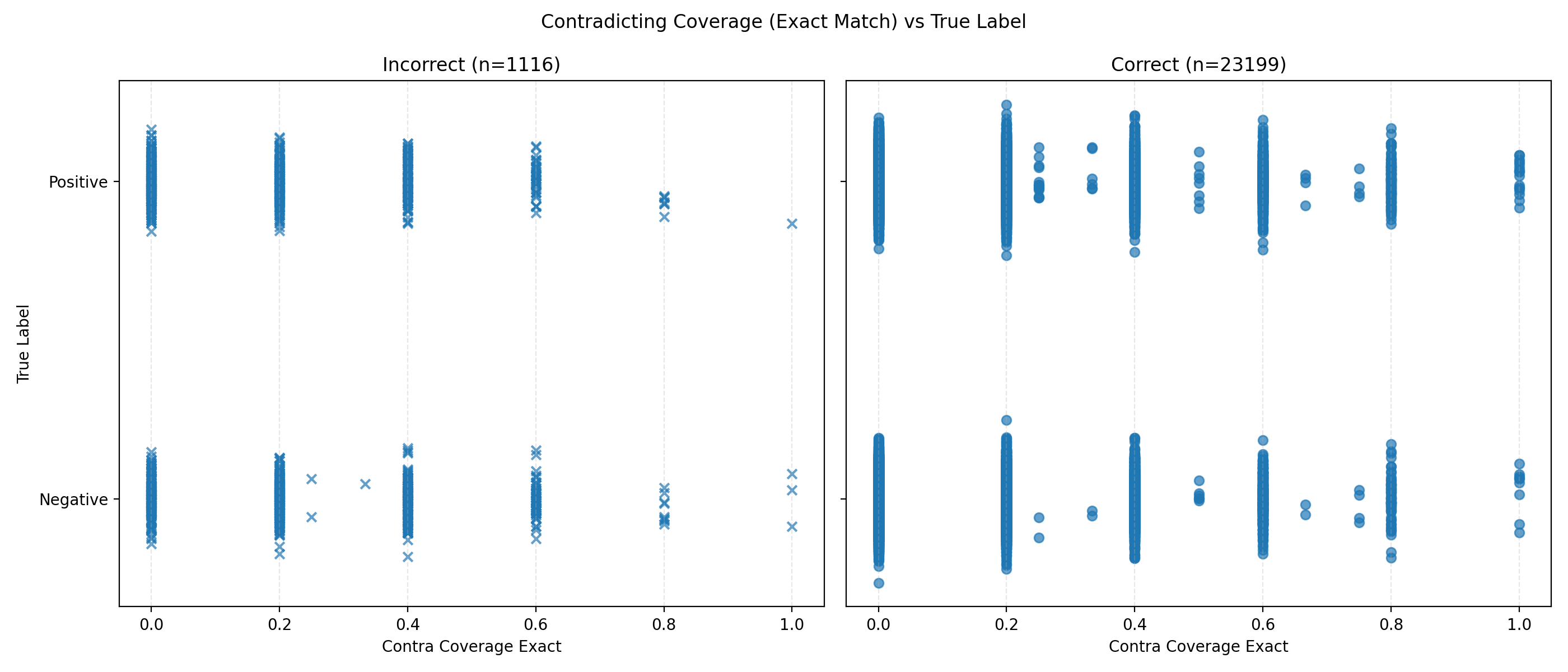}
\caption{Scatter plots of support and contradiction coverage for \textsc{IMDB}.}
\label{fig:imdb-scatter}
\end{figure}

Scatter plots (Figures~\ref{fig:wiki-scatter}–\ref{fig:imdb-scatter}) illustrate this pattern across datasets. Correct predictions cluster in regions with higher support coverage and lower contradiction coverage, while mismatches exhibit greater dispersion and systematically higher contradiction coverage. These visualizations indicate that explanations associated with incorrect predictions more frequently reference lexical cues that oppose the predicted label under the reference classifier.

Table~\ref{tab:coverage-means} reports mean coverage values for supporting and contradicting features at the central feature depth \(k=5\). Across all datasets and both reference model families, explanations accompanying correct predictions consistently exhibit higher coverage of supporting features and lower coverage of contradicting features than explanations associated with incorrect predictions. This qualitative pattern appears consistently across both token-aware and exact matching.

\begin{table*}[t]
\centering
\small
\resizebox{\textwidth}{!}{%
\begin{tabular}{llcccccc}
\toprule
\textbf{Dataset} & \textbf{Model} & \textbf{Condition} & \textbf{Support (Token)} & \textbf{Support (Exact)} & \textbf{Contra (Token)} & \textbf{Contra (Exact)} \\
\midrule

\multirow{4}{*}{\textsc{WikiOntology}}
& \multirow{2}{*}{LogReg}
& Match    & 0.515 & 0.520 & 0.120 & 0.127 \\
&          & Mismatch & 0.302 & 0.302 & 0.211 & 0.217 \\
& \multirow{2}{*}{SVM}
& Match    & 0.516 & 0.522 & 0.134 & 0.139 \\
&          & Mismatch & 0.286 & 0.279 & 0.231 & 0.232 \\

\midrule

\multirow{4}{*}{\textsc{AG News}}
& \multirow{2}{*}{LogReg}
& Match    & 0.534 & 0.533 & 0.154 & 0.157 \\
&          & Mismatch & 0.472 & 0.475 & 0.247 & 0.252 \\
& \multirow{2}{*}{SVM}
& Match    & 0.542 & 0.541 & 0.157 & 0.160 \\
&          & Mismatch & 0.413 & 0.416 & 0.294 & 0.295 \\

\midrule

\multirow{4}{*}{\textsc{IMDB}}
& \multirow{2}{*}{LogReg}
& Match    & 0.277 & 0.321 & 0.130 & 0.139 \\
&          & Mismatch & 0.212 & 0.237 & 0.198 & 0.222 \\
& \multirow{2}{*}{SVM}
& Match    & 0.282 & 0.329 & 0.112 & 0.119 \\
&          & Mismatch & 0.188 & 0.207 & 0.205 & 0.234 \\

\bottomrule
\end{tabular}%
}
\caption{Mean feature coverage at \(k=5\) for logistic regression and SVM reference models. Explanations for correct predictions (\emph{Match}) consistently exhibit higher coverage of supporting features and lower coverage of contradicting features than explanations associated with incorrect predictions (\emph{Mismatch}).}
\label{tab:coverage-means}
\end{table*}

\begin{table}[t]
\centering
\small
\begin{tabular}{lcc}
\toprule
\textbf{Ablation Condition} & \textbf{Accuracy} & \textbf{Error Rate} \\
\midrule
Support Features Removed & 0.8475 & 0.1525 \\
Random Features Removed  & 0.8537 & 0.1463 \\
Contra Features Removed  & \textbf{0.8768} & \textbf{0.1232} \\
\bottomrule
\end{tabular}
\caption{Model accuracy after removing different feature types on AG News. Removing support features slightly reduces performance relative to random removal, while removing contradiction features improves accuracy, suggesting that contradicting cues introduce conflicting signals for the model.}
\label{tab:feature_ablation_accuracy}
\end{table}

\begin{table}[t]
\centering
\small
\begin{tabular}{lc}
\toprule
\textbf{Prediction Transition} & \textbf{Count} \\
\midrule
Random correct $\rightarrow$ Support incorrect & 276 \\
Random incorrect $\rightarrow$ Support correct & 230 \\
Random incorrect $\rightarrow$ Contra correct  & 270 \\
Random correct $\rightarrow$ Contra incorrect  & 98 \\
\bottomrule
\end{tabular}
\caption{Directional prediction changes across ablation conditions. Removing support features breaks correct predictions more often than it fixes errors, while removing contradiction features corrects substantially more errors than it introduces.}
\label{tab:directional_flips}
\end{table}

\subsection{Dataset-Specific Trends}

The magnitude of the support–contra asymmetry varies across datasets. In \textsc{WikiOntology}, the separation between correct and incorrect predictions is most pronounced: explanations accompanying correct predictions show substantially higher support coverage and lower contradiction coverage than those associated with mismatches.

A similar but slightly smaller effect appears in \textsc{AG News}. Correct predictions consistently exhibit higher support coverage and substantially lower contradiction coverage than incorrect predictions across both reference models.

In \textsc{IMDB}, the asymmetry remains present but with smaller margins. Support coverage increases for correct predictions while contradiction coverage rises for mismatches, though the differences are narrower than in the other datasets. This suggests that explanations in sentiment classification rely somewhat less directly on the most salient lexical features identified by the reference classifier.

Despite differences in magnitude, the same qualitative pattern emerges across all datasets: correct predictions align with stronger supporting evidence, whereas incorrect predictions are associated with greater coverage of contradicting cues.

\subsection{Robustness to Retrieval Depth}

To evaluate whether the observed asymmetry depends on the number of classifier features examined per instance, we repeat the analysis with \(k \in \{3,5,7\}\) top-weighted features. We compute bootstrap estimates of the support differential

\[
\Delta_{\text{sup}} = \mu(\text{match}) - \mu(\text{mismatch}),
\]

the contradiction differential

\[
\Delta_{\text{con}} = \mu(\text{match}) - \mu(\text{mismatch}),
\]

and their asymmetry

\[
\Delta^{*} = \Delta_{\text{sup}} - \Delta_{\text{con}}.
\]

Table~\ref{tab:robustness-k} reports the resulting bootstrap estimates and 95\% confidence intervals. Across datasets, the qualitative pattern remains stable as \(k\) varies.

For \textsc{WikiOntology}, support differentials remain strongly positive (\(\approx 0.20\)–\(0.26\)), while contradiction differentials remain negative (\(\approx -0.09\) to \(-0.11\)), yielding large asymmetry values around \(0.29\)–\(0.37\). In \textsc{AG News}, support differentials range from approximately \(0.12\)–\(0.13\) under SVM and \(0.05\)–\(0.06\) under logistic regression, while contradiction differentials remain consistently negative. The \textsc{IMDB} dataset shows moderate asymmetry that gradually decreases as \(k\) increases, reflecting the more diffuse lexical signals associated with sentiment classification.

Importantly, increasing the number of retrieved features slightly attenuates the magnitude of the asymmetry but does not change its direction or statistical significance. This indicates that the support–contra asymmetry is not an artifact of a particular feature cutoff.

\begin{table*}[t]
\centering
\small
\setlength{\tabcolsep}{5pt}
\renewcommand{\arraystretch}{1.15}
\caption{Robustness of coverage differentials across retrieval depth (\(k \in \{3,5,7\}\)).
We report bootstrap estimates of support differential
\(\Delta_{\text{sup}}=\mu(\text{match})-\mu(\text{mismatch})\),
contradiction differential
\(\Delta_{\text{con}}=\mu(\text{match})-\mu(\text{mismatch})\),
and asymmetry
\(\Delta^{*}=\Delta_{\text{sup}}-\Delta_{\text{con}}\).
Results are shown for token and exact matching with 95\% bootstrap confidence intervals.}
\label{tab:robustness-k}
\resizebox{\textwidth}{!}{%
\begin{tabular}{llcccccc}
\toprule
\textbf{Dataset} & \textbf{\(k\)} &
\multicolumn{2}{c}{\textbf{Support \(\Delta_{\text{sup}}\)}} &
\multicolumn{2}{c}{\textbf{Contradict \(\Delta_{\text{con}}\)}} &
\multicolumn{2}{c}{\textbf{Asymmetry \(\Delta^{*}\)}} \\
\cmidrule(lr){3-4}\cmidrule(lr){5-6}\cmidrule(lr){7-8}
 & & Token & Exact & Token & Exact & Token & Exact \\
\midrule

\multirow{3}{*}{\textsc{WikiOntology}}
& 3 & \(0.237[0.148,0.323]\) & \(0.259[0.174,0.340]\) & \(-0.108[-0.198,-0.025]\) & \(-0.107[-0.197,-0.023]\) & \(0.345[0.226,0.470]\) & \(0.366[0.246,0.486]\) \\
& 5 & \(0.213[0.143,0.280]\) & \(0.218[0.147,0.287]\) & \(-0.090[-0.175,-0.013]\) & \(-0.089[-0.173,-0.011]\) & \(0.304[0.201,0.409]\) & \(0.308[0.203,0.414]\) \\
& 7 & \(0.198[0.143,0.251]\) & \(0.198[0.142,0.254]\) & \(-0.093[-0.177,-0.016]\) & \(-0.092[-0.175,-0.014]\) & \(0.291[0.195,0.388]\) & \(0.290[0.194,0.387]\) \\

\midrule

\multirow{3}{*}{\textsc{AG News}}
& 3 & \(0.134[0.115,0.154]\) & \(0.125[0.106,0.145]\) & \(-0.154[-0.173,-0.136]\) & \(-0.154[-0.173,-0.136]\) & \(0.288[0.261,0.316]\) & \(0.280[0.252,0.307]\) \\
& 5 & \(0.129[0.114,0.144]\) & \(0.125[0.110,0.140]\) & \(-0.136[-0.152,-0.121]\) & \(-0.135[-0.151,-0.120]\) & \(0.265[0.243,0.287]\) & \(0.260[0.239,0.282]\) \\
& 7 & \(0.122[0.109,0.134]\) & \(0.118[0.105,0.131]\) & \(-0.124[-0.138,-0.110]\) & \(-0.123[-0.138,-0.109]\) & \(0.246[0.227,0.265]\) & \(0.241[0.222,0.260]\) \\

\midrule

\multirow{3}{*}{\textsc{IMDB}}
& 3 & \(0.076[0.059,0.094]\) & \(0.098[0.080,0.117]\) & \(-0.091[-0.108,-0.075]\) & \(-0.104[-0.121,-0.087]\) & \(0.168[0.143,0.191]\) & \(0.202[0.176,0.227]\) \\
& 5 & \(0.065[0.051,0.079]\) & \(0.084[0.069,0.098]\) & \(-0.069[-0.081,-0.056]\) & \(-0.082[-0.095,-0.070]\) & \(0.134[0.115,0.152]\) & \(0.166[0.146,0.185]\) \\
& 7 & \(0.056[0.044,0.067]\) & \(0.071[0.059,0.083]\) & \(-0.056[-0.066,-0.046]\) & \(-0.067[-0.077,-0.057]\) & \(0.111[0.096,0.126]\) & \(0.138[0.122,0.154]\) \\

\bottomrule
\end{tabular}
}
\end{table*}
\section{Discussion}
\label{discussion}

Our results reveal a consistent support–contra asymmetry in LLM-generated explanations when compared against feature-based evidence from transparent reference classifiers. Across datasets, reference models, and feature depths, explanations accompanying correct predictions reference more supporting lexical evidence and less contradicting evidence than explanations associated with incorrect predictions. This pattern provides a useful empirical perspective on how LLM explanations relate to predictive cues in the input, while also highlighting both the interpretive value and limitations of rationale-based evaluation.

\subsection{Support--Contra Asymmetry in Explanations}
The most striking finding is a systematic asymmetry between supporting and contradicting lexical evidence in LLM explanations. When predictions are correct, explanations show higher overlap with supporting features identified by the reference classifier while rarely mentioning contradicting cues. In contrast, explanations accompanying incorrect predictions exhibit substantially greater coverage of contradicting features. This pattern is visible both in the scatter plots and in the aggregate coverage statistics and bootstrap confidence intervals. Importantly, the effect persists across datasets, feature depths, and reference model choices. The asymmetry statistic remains consistently positive across all experimental configurations, indicating that explanations more frequently reference lexical cues that align with the predictive evidence extracted by the reference classifier when predictions are correct.

One interpretation is that explanations tend to track strong predictive lexical signals present in the input when the model’s prediction aligns with the true label. When predictions are incorrect, however, explanations more often mention cues that the reference classifier associates with competing labels. Under this interpretation, explanation content appears to reflect the distribution of predictive lexical evidence captured by the reference model, rather than functioning purely as unconstrained post-hoc narrative generation.

\subsection{Task Sensitivity of Explanation Alignment}

The magnitude of the asymmetry varies across datasets. In topical classification tasks such as \textsc{WikiOntology} and \textsc{AG News}, explanation--feature alignment is particularly strong. These tasks rely heavily on discriminative lexical signals (e.g., domain-specific terms or entity indicators), making it easier for explanations to reference the same cues identified by the classifier.

In contrast, the effect is more moderate in \textsc{IMDB}. Sentiment classification often relies on a broader distribution of lexical indicators and contextual cues rather than a small set of dominant keywords. As a result, explanations may draw on a wider range of evidence, reducing direct overlap with the classifier's most influential features.

These differences highlight an important consideration for explanation evaluation: the degree of alignment between explanations and feature-based evidence is inherently task dependent. Tasks with strong lexical grounding produce clearer explanation--feature correspondence than tasks where predictive signals are more diffuse.

\subsection{Robustness Across Reference Models}

A key question is whether the observed asymmetry depends on the particular reference model used to extract predictive features. Our experiments address this by comparing two transparent linear classifiers: logistic regression and linear SVM.

Across datasets, both models produce nearly identical qualitative patterns. Supporting coverage increases for correct predictions and contradicting coverage increases for incorrect predictions under both reference models. This consistency suggests that the observed asymmetry reflects a broader relationship between explanations and predictive lexical evidence rather than an artifact of a specific classifier.

Because both models rely on the same TF--IDF representation but differ in their optimization objectives, the agreement between them strengthens the interpretation that explanations are aligning with robust lexical signals in the data.

\subsection{Implications for Explanation--Evidence Alignment}

The observed asymmetry has important implications for explanation evaluation. On one hand, the alignment between explanations and supporting classifier features suggests that explanations often reference lexical cues that are predictive for the task under the external reference models. This indicates that LLM explanations can capture salient evidence present in the input text.

On the other hand, the elevated contradiction coverage observed in incorrect predictions raises concerns about explanation reliability. When predictions are wrong, explanations more often reference cues that oppose the true label under the reference classifier. In such cases, explanations may read as plausible justifications while remaining weakly aligned with predictive evidence associated with correct decisions.

These findings suggest that explanation evaluation should consider not only whether explanations reference predictive evidence, but also how explanation content changes when predictions fail. Examining this shift is important for understanding when explanations are informative and when they may instead reinforce incorrect predictions.

\subsection{Limitations and Future Directions}

While the proposed framework captures systematic patterns of explanation--feature overlap, several limitations remain. First, the reference models used here are linear classifiers based on TF--IDF features. Although these models provide transparent lexical evidence, they cannot capture higher-order compositional or semantic patterns that LLMs may exploit.

Second, the matching procedure relies on lexical overlap between classifier features and explanation text. While token-aware and exact matching capture many correspondences, they cannot fully represent semantic equivalence or deeper paraphrastic relationships.

In addition, because the predictive cues used in this analysis are derived from transparent reference classifiers rather than the LLM itself, the proposed metrics should be interpreted as measuring alignment with an external lexical evidence baseline rather than direct faithfulness to the LLM’s internal decision process.

Finally, the analysis focuses on a limited set of feature depths (\(k \in \{3,5,7\}\)). Although results remain stable across these settings, future work could explore richer attribution methods that capture more distributed forms of predictive evidence.

More broadly, future research could extend this framework by incorporating contextual attribution methods, semantic similarity measures, or human-annotated rationales to obtain a more comprehensive view of explanation--evidence alignment.

\subsection{Summary}

Overall, this study shows that LLM explanations exhibit a robust support--contra asymmetry: explanations for correct predictions emphasize supporting lexical evidence, while explanations for incorrect predictions reference more contradicting cues. This pattern holds across datasets, reference models, and feature depths, suggesting that explanation content is systematically related to predictive evidence present in the input rather than being purely unconstrained post-hoc narrative.

\section{Conclusion}
\label{conclusion}

This study examined how LLM-generated explanations relate to predictive lexical evidence extracted from transparent linear classifiers across three benchmark text classification tasks: \textsc{WikiOntology}, \textsc{AG News}, and \textsc{IMDB}. By comparing explanation content with influential classifier features, we analyzed explanation–evidence alignment in terms of supporting and contradicting feature coverage.

Across datasets, reference models, and feature depths, we observe a consistent empirical pattern that we term \textbf{support–contra asymmetry}. Explanations accompanying correct predictions exhibit higher overlap with supporting classifier features and lower overlap with contradicting features, while explanations associated with incorrect predictions reference substantially more contradicting evidence. This pattern is visible in both descriptive coverage statistics and bootstrap estimates of coverage differentials, and it appears consistently across both logistic regression and linear SVM reference models.

The magnitude of the asymmetry varies across tasks. It is strongest in topical and ontology classification tasks (\textsc{WikiOntology}, \textsc{AG News}), where predictive lexical cues are highly discriminative, and more moderate in sentiment classification (\textsc{IMDB}), where predictive signals are distributed across a broader range of expressions. Nevertheless, the direction of the effect remains stable across all datasets and feature retrieval depths.

Taken together, these findings suggest that LLM explanations often reference lexical cues that align with predictive evidence identified by interpretable reference models when predictions are correct, while explanations accompanying incorrect predictions more frequently highlight cues that oppose the true label under the reference classifier. This pattern indicates that explanation content is systematically related to predictive evidence present in the input rather than being purely unconstrained narrative.

More broadly, our results illustrate how comparing explanations with external sources of predictive evidence can provide useful insight into explanation behavior. The proposed analysis reveals systematic differences in how explanations reference supporting and contradicting cues depending on prediction correctness. Future work could extend this approach by incorporating richer attribution methods, semantic similarity measures, or human-annotated rationales to obtain a more comprehensive picture of explanation–evidence alignment.

Overall, this study demonstrates that LLM explanations exhibit a robust and measurable support–contra asymmetry across tasks and models. Understanding this phenomenon provides a step toward more systematic evaluation of explanation behavior in large language models.

\bibliography{iclr2025_conference}
\nocite{*}
\bibliographystyle{iclr2025_conference}

\end{document}